\RequirePackage{fix-cm}
\RequirePackage{amsmath}
\documentclass[smallextended]{svjour3}       % onecolumn (second format)
\smartqed  % flush right qed marks, e.g. at end of proof

\usepackage[utf8]{inputenc} % allow utf-8 input
\usepackage[T1]{fontenc}    % use 8-bit T1 fonts
\usepackage{hyperref}       % hyperlinks
\usepackage{url}            % simple URL typesetting
\usepackage{booktabs}       % professional-quality tables
\usepackage{amsfonts}      % blackboard math symbols
\usepackage{amsmath}
\usepackage{multirow}
\usepackage{algorithm}
\usepackage{natbib}
\usepackage[english]{babel}
\usepackage[noend]{algorithmic}
\usepackage{nicefrac}       % compact symbols for 1/2, etc.
\usepackage{microtype}      % microtypography
\usepackage{graphicx}
\usepackage{textcomp}
\usepackage{xcolor}
\usepackage[caption=false]{subfig}
\DeclareGraphicsExtensions{.pdf,.jpeg,.jpg,.png}

\newtheorem{assumption}{\textbf{Assumption}}

\newcommand\independent{\protect\mathpalette{\protect\independenT}{\perp}}
\def\independenT#1#2{\mathrel{\rlap{$#1#2$}\mkern2mu{#1#2}}}

\begin{document}

\title{Adversarial Balancing-based Representation Learning for Causal Effect Inference with Observational Data}

\titlerunning{Adversarial Balancing-based Representation Learning for Causal Effect Inference}

\author{Xin Du         \and
        Lei Sun        \and
        Wouter Duivesteijn      \and
        Alexander Nikolaev      \and
        Mykola Pechenizkiy}

%\authorrunning{Short form of author list} % if too long for running head

\institute{X. Du, W. Duivesteijn, M. Pechenizkiy\at
              Eindhoven University of Technology, \\
              \email{\{x.du, w.duivesteijn, m.pechenizkiy\}@tue.nl}           %  \\
%             \emph{Present address:} of F. Author  %  if needed
           \and
           L. Sun, A. Nikolaev \at
              University at Buffalo, \\
              \email{\{leisun, anikolae\}@buffalo.edu}
}

\date{Received: date / Accepted: date}

% \author{Author information scrubbed for blind reviewing}
%\author{%
%  Xin Du\\
%  Eindhoven University of Technology\\
%  \texttt{x.du@tue.nl} \\
%  \And
%  Wouter Duivesteijn \\
%  Eindhoven University of Technology \\
%  \texttt{w.duivesteijn@tue.nl} \\
%  \And
%  Lei Sun \\
%  University at Buffalo \\
%  \texttt{leisun@buffalo.edu} \\
%  \And
%  Alexander Nikolaev \\
%  University at Buffalo \\
%  \texttt{anikolae@buffalo.edu} \\
%  \And
%  Mykola Pechenizkiy \\
%  Eindhoven University of Technology \\
%  \texttt{m.pechenizkiy@tue.nl} \\
%}

\maketitle

\begin{abstract}
  Learning causal effects from observational data greatly benefits a variety of domains such as health care, education and sociology. For instance, one could estimate the impact of a new drug on specific individuals to assist the clinic plan and improve the survival rate. 
  % In this paper, we conduct causal inference with observational studies based on potential outcome framework (PO)~\citep{rubin2005causal}.
  % The central problem for causal effect inference in PO is dealing with the unobserved counterfactuals and treatment selection bias. 
  % The state-of-the-art approaches focus on solving these problems by balancing the treatment and control groups~\citep{sun2016mutual}. 
  In this paper, we focus on studying the problem of estimating Conditional Average Treatment Effect (CATE) from observational data. The challenges for this problem are two-fold: on the one hand, we have to derive a causal estimator to estimate the causal quantity from observational data, where there exists confounding bias; on the other hand, we have to deal with the identification of CATE when the distribution of covariates in treatment and control groups are imbalanced.
  To overcome these challenges, we propose a neural network framework called \textbf{A}dversarial \textbf{B}alancing-based representation learning for \textbf{C}ausal \textbf{E}ffect \textbf{I}nference (\textbf{ABCEI}), based on the recent advances in representation learning. To ensure the identification of CATE, ABCEI uses adversarial learning to balance the distributions of covariates in treatment and control groups in the latent representation space, without any assumption on the form of the treatment selection/assignment function. In addition, during the representation learning and balancing process, highly predictive information from the original covariate space might be lost. ABCEI can tackle this information loss problem by preserving useful information for predicting causal effects under the regularization of a mutual information estimator. 
  %WD: the following sentence should hold for any decent paper, and can hence safely be omitted
  %We conduct various experiments on several synthetic and real-world datasets. 
  The experimental results show that ABCEI is robust against treatment selection bias, and matches/outperforms the state-of-the-art approaches. Our experiments show promising results on several datasets, representing different health care domains among others.
\end{abstract}

\section{Introduction}
Many domains of science require inference of causal effects, including healthcare~\citep{casucci2017estimating}, economics and marketing~\citep{lalonde1986evaluating,smith2005does}, sociology~\citep{morgan2006matching} and education~\citep{zhao2017estimating}. For instance, medical scientists must know whether a new medicine is more beneficial for patients; teachers want to know if their teaching plan can improve the grades of students significantly; economists need to evaluate whether a policy can improve the unemployment rates. Due to the broad application of machine learning model in these domains, properly estimating causal effects is an important task for machine learning research.

The classical method to estimate the causal effects is Randomized Controlled Trials (RCTs)~\citep{autier2007vitamin}, where we have to maintain two statistic identical groups and randomly assign treatments to each individual to observe the outcomes. 
However, RCTs can be time-consuming, expensive, or unethical (e.g.\ for studying the effect of smoking on health condition). Hence, causal effect inference through observational studies are needed~\citep{benson2000comparison}. 
The core issue of causal effect inference from observational data is the identification problem. That is, given a set of assumptions and the non-experimental data, whether it is possible to derive a model that can correctly estimate the strength of causal effect by certain quantities. 

In this paper, our aim is to build a machine learning model that is able to estimate the Conditional Average Treatment Effect (CATE) with observational data. There are several challenges for this task. First, there might be spurious associations between the treatments and outcomes caused by confounding variables: variables that affect both treatment variables and the outcome variables. For example, patients with more personal wealth are in a better position to get new medicines, and at the same time their wealth increases the likelihood that they can survive. 
Due to the existence of confounding bias, it is nearly impossible to build the estimator by directly modeling the relations between treatments and outcomes. Strong ignorability in Potential Outcome framework~\citep{rubin2005causal} provides a way to estimate the causal quantities using the adjustment estimand with statistical quantities. In order to satisfy the ignorability in practical study, people derive methods to match / balance the covariates, e.g. based on mutual information between treatment variables and covariates~\citep{sun2016mutual}, or based on propensity scores~\citep{dehejia2002propensity}. However, these methods are either only feasible for the estimation of Average Treatment Effect (ATE) or Average Treatment effect on the Treated (ATT).
~\cite{pearl2009causality} proposes a criterion based on graphical models to select admissible covariates for ignorability. Throughout this paper, we assume that all the variables in the causal system can be observed and measured, so that the causal effects we are interested in are identifiable from the observational data. This assumption allows us to build causal quantity estimators for each outcome system conditioning on the covariates.

Another challenge for CATE estimation is that in observational study we can only observe the factual outcomes. The counterfactual outcomes can never be observed. When there exists treatment selection bias, the imbalanced distributions of covariates in treatment and control groups would lead to biases for the estimation of CATE due to the generalization error~\citep{swaminathan2015counterfactual}. Various techniques from several studies are proposed to tackle this problem.
~\cite{yao2018representation} propose to use hard samples to preserve local similarity information from covariate space to latent representation space. The hard sample mining process is highly dependent on the propensity score model, which is not robust when the propensity score model is misspecified.~\citep{imai2014covariate,ning2018robust} propose estimators which are robust even when the propensity score model is not correctly specified.~\citep{kallus2018balanced,kallus2018deepmatch,ozery2018adversarial} propose to generate balanced weights for data samples to minimize a selected imbalance measure in covariate space. 
% A logistic regression model is employed to calculate the propensity score. 
~\cite{shalit2017estimating} propose to derive upper bounds on the estimation error by considering both covariate balancing and potential outcomes. Highly predictive information might be lost in the reweighing or balancing processes of these methods.

\begin{figure}[t]
  \centering
  \includegraphics[width=.9\columnwidth]{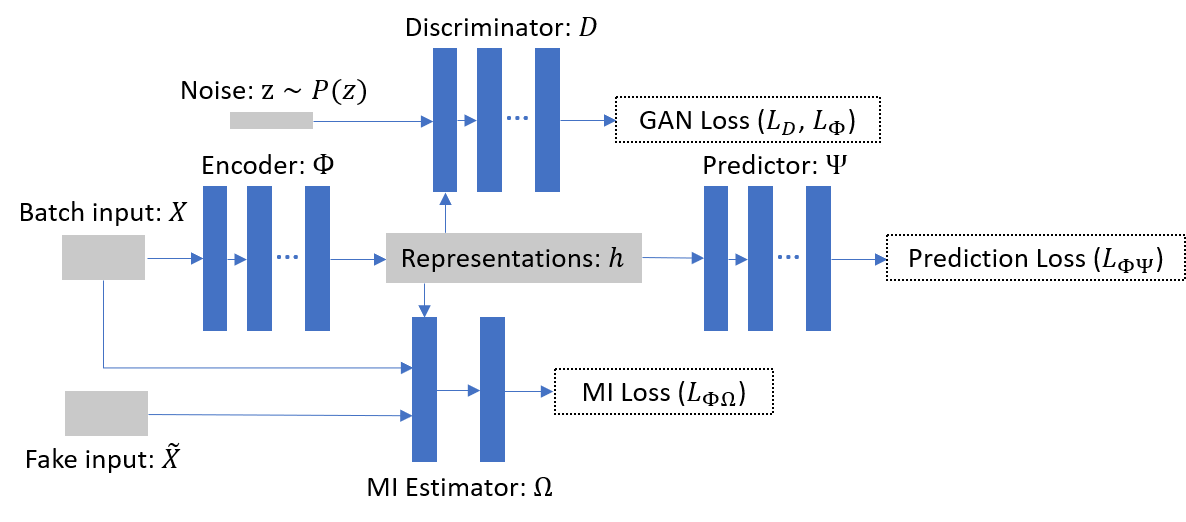}
  \caption{Deep neural network architecture of ABCEI for causal effect inference.}
  \label{fig:arch}
\end{figure}

To address these problems, we propose a framework (cf.\@ Figure \ref{fig:arch}), which generates balanced representations as well as preserving highly predictive information in latent space without using propensity scores. We design a two-player adversarial game, between an encoder that transforms covariates to latent representations and a discriminator which distinguishes representations from control and treatment group. Unlike in the classical GAN framework, here, the `true distribution' (latent representations of the control group\footnote{our method supports representations of either treatment/control group or both as `true distribution'.}) in this game also must be generated by the encoder. On the other hand, to prevent losing useful information during the balancing process, we use a mutual information estimator to constrain the encoder to preserve highly predictive information~\citep{hjelm2018learning}. The outcome data are also considered in this unified framework to specify the causal effect predictor.

Technically, the unified framework encodes the input covariates into a latent representation space, and build estimators to estimate the treatment outcomes with those representations. 
There are three components on top of the encoder in our model: (1) \textbf{mutual information estimation}: an estimator is specified to estimate and maximize the mutual information between representations and covariates; (2) \textbf{adversarial balancing}: the encoder plays an adversarial game with a discriminator, trying to fool the discriminator by minimizing the discrepancies between distributions of representations from the treatment and control group; (3) \textbf{treatment outcome prediction}: a predictor over latent space is employed to estimate the treatment outcomes. By jointly optimizing the three components via back propagation, we can get a robust estimator for the CATE. The overarching architecture of our framework is shown in Figure~\ref{fig:arch}. As a summary, our main contributions are:

% \subsection{Main Contributions}
\begin{enumerate}
\item We propose a novel model: \textbf{A}dversarial \textbf{B}alancing-based representation learning for \textbf{C}ausal \textbf{E}ffect \textbf{I}nference (\textbf{ABCEI}) with  observational data. ABCEI addresses information loss and treatment selection bias by learning highly informative and balanced representations in latent space.
\item A neural network encoder is constrained by a mutual information estimator to minimize the information loss between representations and the input covariates, which preserves highly predictive information for causal effect inference.
\item We employ an adversarial learning method to balance representations between treatment and control groups, which deals with the treatment selection bias problem without any assumption on the form of the treatment selection function, unlike, e.g., the propensity score method.
\item We conduct various experiments on synthetic and real-world datasets. ABCEI outperforms most of the state-of-the-art methods on benchmark datasets. We show that ABCEI is
%WD: what does "decent" mean, here?
% decent and
robust against different experimental settings. By supporting mini-batch, 
%WD: "qualified" is not precise enough.
%ABCEI is qualified for large-scale datasets.
ABCEI can be applied on large-scale datasets.
\end{enumerate}

% \section{Preliminaries}

%WD: I've removed a few empty headers. Those can be informative, but I've kept only those that are informative enough to justify the additional spatial cost
%\subsection{Motivation}
% In this section, we compare our method with recent balancing methods and justify the motivation why we propose the method in this paper.
% In order to properly handle treatment selection bias and counterfactuals, causal effect estimation must solve two central problems: balancing covariates and specifying the outcome model.
%Balancing covariates and specifying outcome model are two central problems for causal effect estimation, in order to account for treatment selection bias and counterfactuals. 

\section{Problem Setup}
Assume an observational dataset $\{X, T, Y\}$, with covariate matrix $X \in \mathbb{R}^{n \times k}$, binary treatment vector $T \in \{0, 1\}^n$, and treatment outcome vector $Y \in \mathbb{R}^n$. Here, $n$ denotes the number of observed units, and $k$ denotes the number of covariates in the dataset.
For each unit $u$, we have $k$ covariates $x_1, \ldots, x_k$, associated with one treatment variable $t \in \{0, 1\}$ and one treatment outcome $y$. According to the Rubin-Neyman causal model~\citep{rubin2005causal}, two potential outcomes $y_0$, $y_1$ exist for treatments $\{0,1\}$, respectively.
% When $t = 1$ is assigned to unit $u$, we say unit $u$ is \emph{treated}, with the outcome $y_1$; otherwise, we say unit $u$ is \emph{untreated} or \emph{control}, with the outcome $y_0$. 
We call $y_{t}$ the \emph{factual outcome}, denoted by $y_f$, and $y_{1-t}$ the \emph{counterfactual outcome}, denoted by $y_{cf}$.
Assuming there is a joint distribution $P(x, t, y_0, y_1)$, we make the following assumptions:
\begin{assumption}[Strong Ignorability]\label{as:as1}
Conditioning on $x$, the potential outcomes $y_0, y_1$ are independent of $t$, which can be stated as: $(y_0, y_1) \independent t | x$.
\end{assumption}

\begin{assumption}[No Interference]\label{as:as3}
The treatment outcome of each individual is not affected by the treatment assignment of other units, which can be formulated as: $Y^u(t^1,\cdots,t^n)=Y^u(t^u)$.
\end{assumption}

\begin{assumption}[Consistency]\label{as:as4}
The potential outcome $y_t$ of each individual is equal to the observed outcome $y$, if the actual treatment received is $T=t$, which can be represented as: $y = y_t$, if $T=t, \forall t$.
\end{assumption}

\begin{assumption}[Positivity]\label{as:as2}
For all sets of covariates and for all treatments, the probability of treatment assignment will always be strictly larger than $0$ and strictly smaller than $1$, which can be expressed as: $0 < P(t|x) < 1, \forall t$ and $\forall x$.
\end{assumption}

Assumption~\ref{as:as1} indicates that all the confounders are observed, i.e., \emph{no unmeasured confounder is present}. Hence by controlling on $X$, we can remove the confounding bias.
Assumption~\ref{as:as2} allows us to estimate the CATE for any $x$ in the covariate space. Under these assumptions, we can formalize the definition of CATE for our task:
\begin{definition}
The Conditional Average Treatment Effect (CATE) for unit $u$ is: $CATE(u) := \mathbb{E}\left[\;y_1\;\middle|\;x^u\;\right] - \mathbb{E}\left[\;y_0\;\middle|\;x^u\;\right]$.
\end{definition}
We can now define the Average Treatment Effect (ATE) and the Average Treatment effect on the Treated (ATT) as: 
\begin{equation*}
ATE := \mathbb{E}\left[\;CATE(u)\;\right] \quad ATT := \mathbb{E}\left[\;CATE(u)\;\middle|\;t=1\;\right].
\end{equation*}
Because the joint distribution $P(x, t, y_0, y_1)$ is unknown, we can only estimate $CATE(u)$ from observational data. A function over the covariate space $\mathcal{X}$ can be defined as $f:\mathcal{X} \times \{0, 1\} \rightarrow \mathcal{Y}$. The estimate of ${CATE}(u)$ can now be defined:
\begin{definition}
Given an observational dataset $\{X, T, Y\}$ and a function $f$, for unit $u$, the estimate of $CATE(u)$ is: 
\begin{equation*}
\widehat{CATE}(u) = f(x^u, 1) - f(x^u, 0).
\end{equation*}
\end{definition}

In order to properly accomplish the task of CATE estimation, we need to find an optimal function over the covariate space for both systems ($t=1$ and $t=0$). 

% For the observational dataset, we only know the factual outcomes. If we directly apply the classical supervised learning framework, we will get a biased model, which will suffer from high generalization error~\citep{swaminathan2015batch}.

\section{Proposed Method}
In order to overcome the challenges in CATE estimation, we build our model on recent advances in representation learning. We propose to define a function $\Phi: \mathcal{X} \rightarrow \mathcal{H}$, and a function $\Psi : \mathcal{H} \rightarrow \mathcal{Y}$. Then we have $\widehat{Y_T} = f(X,T) = \Psi(\Phi(X),T) = \Psi(h,T)$. Instead of directly estimating the treatment outcome conditioned on covariates, we firstly use an encoder to learn latent representations of covariates. We simultaneously learn latent representations and estimate the treatment outcome. However, the function $f$ would still suffer from information loss and treatment selection bias, unless we constrain the encoder $\Phi$ to learn balanced representations while preserving useful information.

\begin{figure}[t]
\centering
  \includegraphics[width=.6\textwidth]{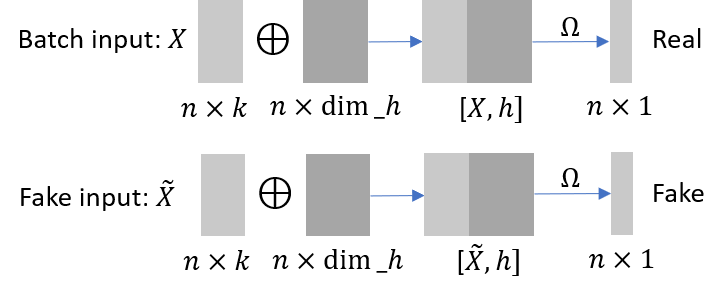}
  \caption{MI estimator between covariates and latent representations.}
  \label{fig:mine}
\end{figure}

\subsection{Mutual Information Estimation}
Consider the information loss when transforming covariates into latent space.  
The non-linear statistical dependencies between variables can be acquired by mutual information (MI)~\citep{kinney2014equitability}.
Thus we use MI between latent representations and original covariates as a measure to account for information loss: 
\begin{equation*}\label{eq:MIcal}
I(X;h) = \int_{\mathcal{X}}\int_{\mathcal{H}}P(x,h)\log\left(\frac{P(x,h)}{P(x)P(h)}\right)\text{d}h\;\text{d}x.
\end{equation*}
We denote the joint distribution between covariates and representations by $\mathbb{P}_{Xh}$ and the product of marginals by $\mathbb{P}_{X} \otimes \mathbb{P}_{h}$.
From the viewpoint of Shannon information theory, mutual information can be represented as Kullback-Leibler (KL) divergence:
\begin{equation*}
I(X;h) := H(X) - H(X|h) := D_{KL}(\mathbb{P}_{Xh}||\mathbb{P}_{X} \otimes \mathbb{P}_{h}),
\end{equation*}
% According to Equation (\ref{eq:MIcal}), i
It is hard to compute MI in continuous and high-dimensional spaces, but one can capture a lower bound of MI with the Donsker-Varadhan representation of KL-divergence~\citep{donsker1983asymptotic}: 
\begin{theorem}[Donsker-Varadhan]
\begin{equation*}
\begin{split}
D_{KL}(\mathbb{P}_{Xh}||\mathbb{P}_{X} \otimes \mathbb{P}_{h}) = \sup_{\Omega \in \mathcal{C}} \mathbb{E}_{\mathbb{P}_{Xh}}[\Omega(x,h)] - \log\mathbb{E}_{\mathbb{P}_{X} \otimes \mathbb{P}_{h}}\left[e^{\Omega(x,h)}\right].
\end{split}
\end{equation*}
\end{theorem}
Here, $\mathcal{C}$ denotes the set of unconstrained functions $\Omega$.

\begin{proof}
Given a fixed function $\Omega$, we can define distribution $G$ by:
\begin{equation*}
\mathrm{d}G = \frac{e^{\Omega(Z)}\mathrm{d}Q}{\int_{\mathcal{Z}}e^{\Omega(Z)}\mathrm{d}Q}
\end{equation*}
Equivalently, we have:
\begin{align*}
\begin{split}
\mathrm{d}G = e^{(\Omega(Z) - S)}\mathrm{d}Q\,\,,
\end{split}
\begin{split}
S = \log \mathbb{E}_{Q}\left[e^{\Omega(Z)}\right]
\end{split}
\end{align*}
Then by construction, we have:
\begin{equation*}
\begin{split}
&\mathbb{E}_{P}[\Omega(Z)] - \log \mathbb{E}_{Q}\left[e^{\Omega(Z)}\right] \\&
= \mathbb{E}_{P}[\Omega(Z)] - S \\&
= \mathbb{E}_{P}\left[\log\frac{\mathrm{d}G}{\mathrm{d}Q}\right] \\&
= \mathbb{E}_{P}\left[\log\frac{\mathrm{d}P\mathrm{d}G}{\mathrm{d}Q\mathrm{d}P}\right] \\&
= \mathbb{E}_{P}\left[\log\frac{\mathrm{d}P}{\mathrm{d}Q} - \log\frac{\mathrm{d}P}{\mathrm{d}G}\right] \\&
= D_{KL}(P||Q) - D_{KL}(P||G) \\&
\leq D_{KL}(P||Q)
\end{split}
\end{equation*}
When distribution $G$ is equal to $P$, this bound is tight. 
\end{proof}

Inspired by Mutual Information Neural Estimation (MINE)~\citep{pmlr-v80-belghazi18a}, we propose to establish a neural network estimator for MI. Specifically, let $\Omega$ be a function: $\mathcal{X} \times \mathcal{H} \rightarrow \mathbb{R}$ parametrized by a deep neural network, we have:
\begin{equation}\label{eq:EMI}
\begin{split}
&I(X;h) := D_{KL}\left(\mathbb{P}_{Xh}\middle|\middle|\mathbb{P}_{X} \otimes \mathbb{P}_{h}\right) \geq \hat{I}_{\Omega}(X;h)\\
&:= \mathbb{E}_{\mathbb{P}_{Xh}}[\Omega(x,h)] - \log\mathbb{E}_{\mathbb{P}_{X} \otimes \mathbb{P}_{h}}\left[e^{\Omega(x,h)}\right].
\end{split}
\end{equation}
By distinguishing the joint distribution and the product of marginals, the estimator $\Omega$ approximates the MI with arbitrary precision. In practice, as shown in Figure~\ref{fig:mine}, we concatenate the input covariates $X$ with representations $h$ one by one to create positive samples (as samples from the true joint distribution). Then, we randomly shuffle $X$ on the batch axis to create fake input covariates $\tilde{X}$. Representations $h$ are concatenated with fake input $\tilde{X}$ to create negative samples (as samples from the product of marginals).
From Equation (\ref{eq:EMI}) we can derive the loss function for the MI estimator:
\begin{equation*}
L_{\Phi \Omega} = -\mathbb{E}_{x \sim X}\left[\Omega\left(x,h\right)\right] + \log\mathbb{E}_{x \sim \tilde{X}}\left[e^{\Omega\left(x,h\right)}\right].
\end{equation*}
Information loss can be diminished by simultaneously optimizing the encoder $\Phi$ and the MI estimator $\Omega$ to minimize $L_{\Phi \Omega}$ iteratively via gradient descent.

\subsection{Adversarial Balancing}
The representations of treatment and control groups are denoted by $h(t=1)$ and $h(t=0)$, corresponding to the input covariate groups $X(t=1)$ and $X(t=0)$. The discrepancy between distributions of the treatment and control groups is an urgent problem in need of a solution. To decrease this discrepancy, we propose an adversarial learning method to constrain the encoder to learn treatment and control representations that are balanced distributions. We build an adversarial game between a discriminator $D$ and the encoder $\Phi$, inspired by the framework of Generative Adversarial Networks (GAN)~\citep{goodfellow2014generative}. In the classical GAN framework, a source of noise is mapped to a generated image by a generator. A discriminator is trained to distinguish whether an input sample is from true or synthetic image distribution generated by the generator. The aim of classical GAN is training a reliable discriminator to distinguish fake and real images, and using the discriminator to train a generator to generate images by fooling the discriminator. 
% However, sometimes the training process of classical GAN is unstable.

In our adversarial game: (1) we draw a noise vector $z \sim P(z)$ which has the same length as the latent representations, where $P(z)$ can be a spherical Gaussian distribution or a Uniform distribution; (2) we separate representation by treatment assignment, and form two distributions: $P_{h(t=1)}$ and $P_{h(t=0)}$; (3) we train a discriminator $D$ to distinguish concatenated vectors from treatment and control group ($[z, h(t=1)]$ and $[z, h(t=0)]$); (4) we optimize the encoder $\Phi$ to generate balanced representations to fool the discriminator.

\begin{figure*}[t]
  \centering
  \includegraphics[width=.9\columnwidth]{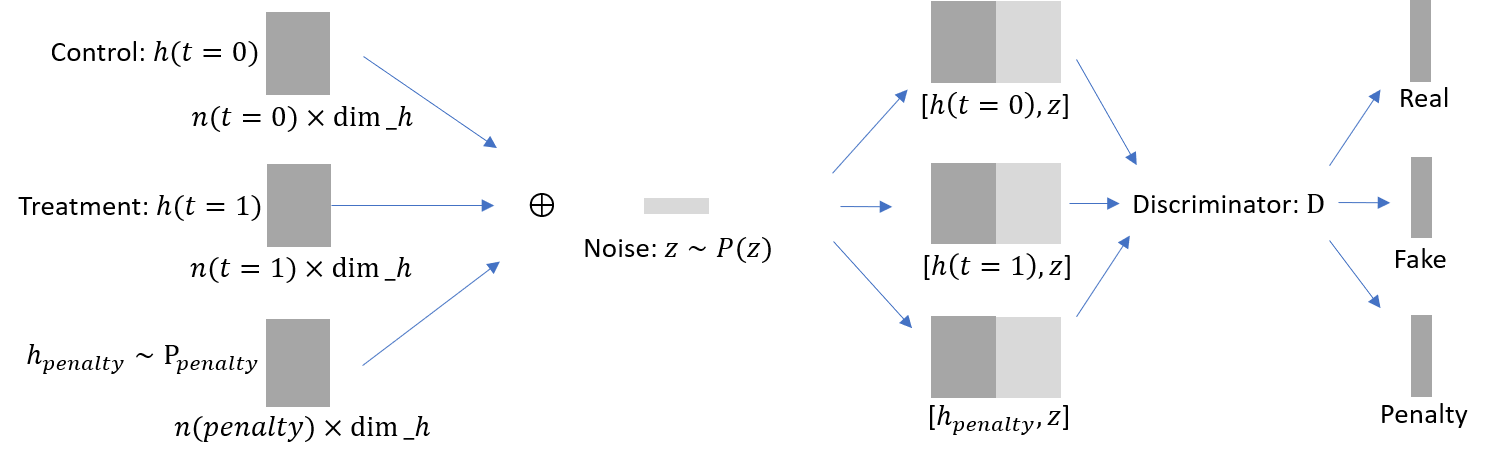}
  \caption{Adversarial learning structure for representation balancing.}
  \label{fig:ganbal}
\end{figure*}

According to the architecture of ABCEI, the encoder is associated with the MI estimator $\Omega$, treatment outcome predictor $\Psi$ and adversarial discriminator $D$. This means that the training process is iteratively adjusting each of the components. The instability of GAN training will become serious in this context. To stabilize the training of GAN, we propose to use the framework of Wasserstein GAN with gradient penalty~\citep{gulrajani2017improved}. By removing the sigmoid layer and applying the gradient penalty to the data between the distributions of treatment and control groups, we can find a function $D$ which satisfies the $\text{1-Lipschitz}$ inequality: 
\begin{equation*}
\left|\left|D\left(x^1\right) - D\left(x^2\right)\right|\right| \leq \left|\left|x^1 - x^2\right|\right|.
\end{equation*}
We can write down the form of our adversarial game:
\begin{equation*}
\begin{split}
\min_{\Phi}\max_{D} \mathbb{E}_{h \sim P_{h(t=0)}}[D([z,h])] - \mathbb{E}_{h \sim P_{h(t=1)}}[D([z,h])] - \\\beta \, \mathbb{E}_{h \sim P_{\text{penalty}}}\left[(||\nabla_{[z,h]}D([z,h])||_2 - 1)^2\right],
\end{split}
\end{equation*}
where $P_{\text{penalty}}$ is the distribution acquired by uniformly sampling along the straight lines between pairs of samples from $P_{h(t=0)}$ and $P_{h(t=1)}$.
The adversarial learning process is in Figure~\ref{fig:ganbal}.

This ensures the encoder $\Phi$ to be smoothly trained to generate balanced representations. 
We can write down the training objective for discriminator and encoder, respectively:
\begin{align*}
%\begin{split}
L_D =& -\mathbb{E}_{h \sim P_{h(t=0)}}[D([z,h])] + \mathbb{E}_{h \sim P_{h(t=1)}}[D([z,h])] \\
&+ \beta \, \mathbb{E}_{h \sim P_{\text{penalty}}}\left[(||\nabla_{[z,h]}D([z,h])||_2 - 1)^2\right],\\
%\end{split}\\
%\begin{split}
L_{\Phi} =& \mathbb{E}_{h \sim P_{h(t=0)}}[D([z,h])] - \mathbb{E}_{h \sim P_{h(t=1)}}[D([z,h])]. 
%\end{split}
\end{align*}

\subsection{Treatment Outcome Prediction}
The final step for CATE estimation is to predict the treatment outcomes with learned representations. We establish a neural network predictor, which takes latent representations and treatment assignments of units as the input, to conduct outcome prediction: $\widehat{y_t} = \Psi(h,t)$. We can write down the loss function of the training objective as:
\begin{equation*}
L_{\Phi \Psi} = \mathbb{E}_{(h,t,y_{t}) \sim \{h,T,Y_T\}}\left[\left(\Psi\left(h,t\right) - y_{t}\right)^2\right] + \lambda \, R(\Psi).
\end{equation*}
Here, $R$ is a regularization on $\Psi$ for the model complexity.

\subsection{Learning Optimization}
W.r.t.\@ the architecture in Figure~\ref{fig:arch}, we minimize $L_{\Phi \Omega}$, $L_{\Phi}$, and $L_{\Phi \Psi}$, respectively, to iteratively optimize parameters in the global model. The optimization steps are handled with the stochastic method Adam~\citep{kingma2014adam}, training the model within Algorithm~\ref{al:alg1}. Optimization details and computational complexity analysis are given in the supplementary material. 

\begin{figure}[htb]
\centering
% \resizebox{0.85\textwidth}{!}
{
\begin{minipage}{\textwidth}
\begin{algorithm}[H]
\caption{ABCEI}
\begin{algorithmic}
  \STATE Input: Observational dataset $\{X, T, Y\}$; loss function $L_{\Phi \Omega}$, $L_{\Phi}$ and $L_{\Phi \Psi}$, $L_D$; Neural Networks $\Phi$, $\Omega$, $D$, $\Psi$; parameters $\Theta_{\Phi}$, $\Theta_{\Omega}$, $\Theta_D$, $\Theta_{\Psi}$ 
  \REPEAT
  \STATE Draw mini-batch $\{X_b, T_b, Y_b\} \subset \{X, T, Y\}$
  \STATE Compute representations $h = \Phi(X_b)$
  \STATE Draw fake input $\tilde{X}_b \sim \mathbb{\tilde{P}}$
  \STATE Draw noise $z \sim \mathcal{N}(0,I)$
  \STATE Set $\Theta_{\Phi}$, $\Theta_{\Omega}$ $\leftarrow$ $\text{Adam}(L_{\Phi \Omega}(X_b,\tilde{X}_b,h),\Theta_{\Phi},\Theta_{\Omega})$
  \FOR{$i=1$ to $3$}
  \STATE Set $\Theta_D$ $\leftarrow$ $\text{Adam}(L_D(h,z,T_b),\Theta_D)$
  \ENDFOR 
  \STATE Set $\Theta_{\Phi}$ $\leftarrow$ $\text{Adam}(L_{\Phi}(h,z,T_b),\Theta_{\Phi})$
  \STATE Set $\Theta_{\Phi}$, $\Theta_{\Psi}$ $\leftarrow$ $\text{Adam}(L_{\Phi \Psi}(h,T_b,Y_b),\Theta_{\Phi}, \Theta_{\Psi})$
  \UNTIL{convergence}
\end{algorithmic}
\label{al:alg1}
\end{algorithm}
\end{minipage}
}
\end{figure}

\section{Experiments}
Due to the lack of counterfactual treatment outcomes in observational data, it is difficult to validate and test the performance of causal effect inference methods. 
In this paper, we adopt two ways to construct the datasets that are available for validating and testing the performance of causal inference methods: the one is to use simulated or semi-simulated treatment outcomes, e.g., dataset IHDP~\citep{hill2011bayesian}; the other is to use RCT datasets and add a non-randomized component to generate imbalanced datasets, e.g., dataset Jobs~\citep{lalonde1986evaluating,smith2005does}. We employ five benchmark datasets: IHDP, Jobs, Twins~\citep{louizos2017causal}, ACIC~\citep{dorie2019automated} and MIMICiii~\citep{johnson2016mimic,johnson2019mimic}. For IHDP, Jobs, Twins, ACIC, and MIMICIII, the experimental results are averaged over $1000,100,100,7700,100$ train/validation/test sets respectively with split sizes $60\%/30\%/10\%$. The implementation of our method is based on Python and Tensorflow~\citep{abadi2016tensorflow}. 
All the experiments in this paper are conducted on a cluster with 1x Intel Xeon E5 2.2GHz CPU, 4x Nvidia Tesla V100 GPU and 256GB RAM. The source code of our algorithms is available on GitHub\footnote{\url{https://github.com/octeufer/Adversarial-Balancing-based-representation-learning-for-Causal-Effect-Inference}}.

\subsection{Details of Datasets}

\paragraph{\textbf{IHDP}} The \emph{Infant Health and Development Program} (IHDP) studies the impact of specialist home visits on future cognitive test scores. Covariates in the semi-simulated dataset are collected from a real-world randomized experiment. The treatment selection bias is created by removing a subset of the treatment group. We use the setting `A' in~\citep{dorie2016npci} to simulate treatment outcomes. This dataset includes $747$ units ($608$ control and $139$ treated) with $25$ covariates associated with each unit.

\paragraph{\textbf{Jobs}} The \emph{Jobs} dataset~\citep{lalonde1986evaluating,smith2005does} studies the effect of job training on the employment status. It consists of a non-randomized component from observational studies and a randomized component based on the National Supported Work program. The randomized component includes $722$ units ($425$ control and $297$ treated) with seven covariates, and the non-randomized component (PSID comparison group) includes $2490$ control units.

\paragraph{\textbf{Twins}} The \emph{Twins} dataset is created based on the ``Linked Birth / Infant Death Cohort Data'' by NBER~\footnote{\url{https://nber.org/data/linked-birth-infant-death-data-vital-statistics-data.html}}. Inspired by~\citep{almond2005costs}, we employ a matching algorithm to select twin births in the USA between 1989-1991. By doing this, we get units associated with $43$ covariates including education, age, race of parents, birth place, marital status of mother, the month in which pregnancy prenatal care began, total number of prenatal visits and other variables indicating demographic and health conditions. We only select twins that have the same gender who both weigh less than $2000g$. For the treatment variable, we use $t=0$ indicating the lighter twin and $t=1$ indicating the heavier twin. We take the mortality of each twin in their first year of life as the treatment outcome, inspired by~\citep{louizos2017causal}. Finally, we have a dataset consisting of 12,828 pairs of twins whose mortality rate is $19.02\%$ for the lighter twin and $16.54\%$ for the heavier twin. Hence, we have observational treatment outcomes for both treatments. In order to simulate the selection bias, we selectively choose one of the twins to observe with regard to the covariates associated with each unit as follows: $t|x \sim \text{Bernoulli}(\sigma(w^T x + n))$, where $w^T \sim \mathcal{N}(0,0.1\cdot I)$ and $n \sim \mathcal{N}(1,0.1)$. 

\paragraph{\textbf{ACIC}} The \emph{Atlantic Causal Inference Conference} (ACIC)~\citep{dorie2019automated} is derived from real-world data with 4802 observations using 58 covariates. There are 77 datasets which are simulated with different treatment selection and outcome functions. Each dataset is generated with 100 random replications independently. In this benchmark, different settings like degrees of non-linearity, treatment selection bias and magnitude of treatment outcome are considered.

\paragraph{\textbf{MIMICIII}} This benchmark is created based on \emph{MIMIC-III}, a database comprising deidentified healthcare data associated with patients data in critical care units. We select patient samples with their demographic information as well as the first observed various laboratory measurements by chemistry or hematology. After filtering samples with missing values, the benchmark consists of 7413 samples with 25 covariates. We would like to investigate the effect of prescription amount in the first day of critical care unit on the length of stay in the ICU. Here we choose binary treatment where 0 represents small amount of prescription and 1 represents large amount of prescription. The treatment outcomes are simulated by $y|x,t \sim (w^Tx + \beta t + n)$, where $n \sim \mathcal{N}(0, 1)$, $w \sim \mathcal{N}(0^{25}, 0.5 \cdot (\Sigma + \Sigma^T))$, and $\Sigma \sim \mathcal{U}((-1,1)^{25 \times 25})$. The treatment assignments are simulated by $t|x \sim Bernoulli(\sigma(s^Tx + m))$, where $m \sim \mathcal{N}(0, 0.1)$ and $s \sim \mathcal{N}(0^{25},0.1\cdot I)$.

\subsection{Evaluation Metrics}
Since the ground truth CATE for the IHDP dataset and MIMICIII benchmark is known, we can employ Precision in Estimation of Heterogeneous Effect (PEHE)~\citep{hill2011bayesian}, as the evaluation metric of CATE estimation: 
\begin{equation*}
\epsilon_{PEHE} = \frac{1}{n}\sum^n_{u=1}((\mathbb{E}[y_1|x^u] - \mathbb{E}[y_0|x^u])-(f(x^u,1)-f(x^u,0)))^2.
\end{equation*}
Subsequently, we can evaluate the precision of ATE estimation based on estimated CATE.  For the Jobs dataset, because we only know parts of the ground truth (the randomized component), we cannot evaluate the performance of ATE estimation. Following~\citep{shalit2017estimating}, we evaluate the precision of ATT estimation and policy risk estimation, where 
\begin{equation*}
R_{pol}(\pi) = 1 - \left[\mathbb{E}\left(y_1\middle|\pi\left(x^u\right)=1\right) \cdot P(\pi = 1) + \mathbb{E}\left(y_0\middle|\pi\left(x^u\right)=0\right) \cdot P(\pi = 0)\right].
\end{equation*}
In this paper, we consider $\pi(x^u) = 1$ when $f(x^u,1) - f(x^u,0) > 0$. 
For the Twins dataset, because we only know the observed treatment outcome for each unit, we follow~\citep{louizos2017causal} using area under ROC curve (AUC) as the evaluation metric. For ACIC dataset, we follow~\citep{ozery2018adversarial} to use RMSE ATE as performance metric.

% \subsection{Evaluation Metrics}
% Since the ground truth ITE for the IHDP dataset is known, we can employ Precision in Estimation of Heterogeneous Effect (PEHE)~\cite{hill2011bayesian}, as the evaluation metric of ITE estimation: $\epsilon_{PEHE} = \frac{1}{n}\sum^n_{u=1}((\mathbb{E}[y_1|x^u] - \mathbb{E}[y_0|x^u])-(f(x^u,1)-f(x^u,0)))^2$.
% Subsequently, we can evaluate the precision of ATE estimation based on estimated ITE.  For the Jobs dataset, because we only know parts of the ground truth (the randomized component), we cannot evaluate the performance of ATE estimation. Following~\cite{shalit2017estimating}, we evaluate the precision of ATT estimation and policy risk estimation, where $R_{pol}(\pi) = 1 - \left[\mathbb{E}\left(y_1\middle|\pi\left(x^u\right)=1\right) \cdot P(\pi = 1) + \mathbb{E}\left(y_0\middle|\pi\left(x^u\right)=0\right) \cdot P(\pi = 0)\right]$.
% In this paper, we consider $\pi(x^u) = 1$ when $f(x^u,1) - f(x^u,0) > 0$. 
% For the Twins dataset, because we only know the observed treatment outcome for each unit, we follow~\cite{louizos2017causal} using area under ROC curve (AUC) as the evaluation metric. For ACIC dataset, we follow~\cite{ozery2018adversarial} to use RMSE ATE as performance metric.

\subsection{Baseline Methods}
We compare with the following baselines: least square regression using treatment as a feature (\textbf{OLS/$\mathbf{LR_1}$}); separate least square regressions for each treatment (\textbf{OLS/$\mathbf{LR_2}$}); balancing linear regression (\textbf{BLR}) and balancing neural network (\textbf{BNN})~\cite{johansson2016learning}; $k$-nearest neighbor (\textbf{k-NN})~\cite{crump2008nonparametric}; Bayesian additive regression trees (\textbf{BART})~\cite{sparapani2016nonparametric}; random forests (\textbf{RF})~\cite{breiman2001random}; causal forests (\textbf{CF})~\cite{wager2017estimation}; treatment-agnostic representation networks (\textbf{TARNet}) and counterfactual regression with Wasserstein distance (\textbf{CFR-Wass})~\cite{shalit2017estimating}; causal effect variational autoencoders (\textbf{CEVAE})~\cite{louizos2017causal}; local similarity preserved individual treatment effect (\textbf{SITE})~\cite{yao2018representation}. MMD measure using RBF kernel (\textbf{MMD-V1, MMD-V2})~\cite{kallus2018deepmatch,kallus2018balanced}. Adversarial balancing with cross-validation procedure (\textbf{ADV-LR/SVM/MLP})~\cite{ozery2018adversarial}. We show the quantitative comparison between our method and the state-of-the-art baselines. 
Experimental results (in-sample and out-of-sample) on IHDP, Jobs and Twins datasets are reported. Specifically, we use $\text{ABCEI}^*$ to represent our model without the mutual information estimation component, and $\text{ABCEI}^{**}$ to represent our model without the adversarial learning component.

\begin{table}[t]
\caption{In-sample and out-of-sample results with mean and standard errors on the IHDP and Jobs dataset (lower = better).}
\label{exp:ihdp}
\centering
% \begin{tabular}{l|c|c|cc}
% \resizebox{0.98\textwidth}{!}
{%
\begin{tabular}{l|l|l|l|l}
{\multirow{3}{*}{Methods}} & \multicolumn{4}{c}{IHDP}                                      \\ \cline{2-5} 
{} & \multicolumn{2}{c|}{In-sample} & \multicolumn{2}{c}{Out-sample} \\ \cline{2-5}
% \multirow{2}{*}{Methods} & \multicolumn{2}{c|}{In-sample} & \multicolumn{2}{c}{Out-sample}                     \\ \cline{2-5} 
{} & $\sqrt{\epsilon_{PEHE}}$ & $\epsilon_{ATE}$ & $\sqrt{\epsilon_{PEHE}}$ & $\epsilon_{ATE}$    \\ \hline
OLS/$LR_1$ & $5.8\phantom{0} \pm .3$  & $.73 \pm .04$ & $5.8\phantom{0} \pm .3$  & $.94 \pm .06$  \\
OLS/$LR_2$ & $2.4\phantom{0} \pm .1$  & $.14 \pm .01$ & $2.5\phantom{0} \pm .1$  & $.31 \pm .02$  \\
BLR        & $5.8\phantom{0} \pm .3$  & $.72 \pm .04$ & $5.8\phantom{0} \pm .3$  & $.93 \pm .05$  \\
BART       & $2.1\phantom{0} \pm .1$  & $.23 \pm .01$ & $2.3\phantom{0} \pm .1$  & $.34 \pm .02$  \\ \hline
k-NN       & $2.1\phantom{0} \pm .1$  & $.14 \pm .01$ & $4.1\phantom{0} \pm .2$  & $.79 \pm .05$  \\
RF         & $4.2\phantom{0} \pm .2$  & $.73 \pm .05$ & $6.6\phantom{0} \pm .3$  & $.96 \pm .06$  \\
CF         & $3.8\phantom{0} \pm .2$  & $.18 \pm .01$ & $3.8\phantom{0} \pm .2$  & $.40 \pm .03$  \\ \hline
BNN        & $2.2\phantom{0} \pm .1$  & $.37 \pm .03$ & $2.1\phantom{0} \pm .1$  & $.42 \pm .03$  \\
TARNet     & $\phantom{0}.88 \pm .0$  & $.26 \pm .01$ & $\phantom{0}.95 \pm .0$  & $.28 \pm .01$  \\
CFR-Wass   & $\phantom{0}.71 \pm .0$  & $.25 \pm .01$ & $\phantom{0}.76 \pm .0$  & $.27 \pm .01$  \\
CEVAE      & $2.7\phantom{0} \pm .1$  & $.34 \pm .01$ & $2.6\phantom{0} \pm .1$  & $.46 \pm .02$  \\
SITE       & $\mathbf{\phantom{0}.69 \pm .0}$  & $.22 \pm .01$  & $\phantom{0}.75 \pm .0$   & $.24 \pm .01$ \\ \hline
$\text{ABCEI}^*$  & $\phantom{0}.74 \pm .0$ & $.12 \pm .01$  & $\phantom{0}.78 \pm .0$  &  $.11 \pm .01$  \\
$\text{ABCEI}^{**}$  & $\phantom{0}.81 \pm .1$ & $.18 \pm .03$  & $\phantom{0}.89 \pm .1$ &     $.16 \pm .02$ \\
$\text{ABCEI}$      & $\phantom{0}.71 \pm .0$                 & $\mathbf{.09 \pm .01}$      & $\mathbf{\phantom{0}.73 \pm .0}$          &     $\mathbf{.09 \pm .01}$ \\ \hline
{\multirow{3}{*}{Methods}} & \multicolumn{4}{c}{Jobs}   \\ \cline{2-5} 
{} & \multicolumn{2}{c|}{In-sample} & \multicolumn{2}{c}{Out-sample} \\ \cline{2-5}
% \multirow{2}{*}{Methods} & \multicolumn{2}{c|}{In-sample} & \multicolumn{2}{c}{Out-sample}                     \\ \cline{2-5} 
{} & $R_{pol}$ & $\epsilon_{ATT}$ & $R_{pol}$ & $\epsilon_{ATT}$ \\ \hline
OLS/$LR_1$ &  $.22 \pm .0$ & $\mathbf{.01 \pm .00}$  & \multicolumn{1}{c|}{$.23 \pm .0$}          &     $.08 \pm .04$      \\
OLS/$LR_2$ &  $.21 \pm .0$ & $.01 \pm .01$ & $.24 \pm .0$  &  $.08 \pm .03$     \\
BLR        &  $.22 \pm .0$ & $.01 \pm .01$ & $.25 \pm .0$  &  $.08 \pm .03$      \\
BART       &  $.23 \pm .0$ & $.02 \pm .00$ & $.25 \pm .0$  &  $.08 \pm .03$         \\ \hline
k-NN       &  $.23 \pm .0$ & $.02 \pm .01$ & $.26 \pm .0$  &  $.13 \pm .05$     \\
RF         &  $.23 \pm .0$ & $.03 \pm .01$ & $.28 \pm .0$  &  $.09 \pm .04$     \\
CF         &  $.19 \pm .0$ & $.03 \pm .01$ & $.20 \pm .0$  &  $.07 \pm .03$    \\ \hline
BNN        &  $.20 \pm .0$ & $.04 \pm .01$ & $.24 \pm .0$  &  $.09 \pm .04$       \\
TARNet     &  $.17 \pm .0$ & $.05 \pm .02$ & $.21 \pm .0$  &  $.11 \pm .04$    \\
CFR-Wass   &  $.17 \pm .0$ & $.04 \pm .01$ & $.21 \pm .0$  &  $.08 \pm .03$     \\
CEVAE      &  $.15 \pm .0$ & $.02 \pm .01$ & $.26 \pm .1$  &  $.03 \pm .01$      \\
SITE       &  $.17 \pm .0$ & $.04 \pm .01$ & $.21 \pm .0$ & $.09 \pm .03$      \\ \hline
$\text{ABCEI}^*$  & $.14 \pm .0$        & $.04 \pm .01$        & $.18 \pm .0$          &     $.04 \pm .01$ \\
$\text{ABCEI}^{**}$ & $.15 \pm .0$        & $.05 \pm .01$        & $.19 \pm .0$          &     $.04 \pm .01$  \\
$\text{ABCEI}$   & $\mathbf{.13 \pm .0}$        & $.02 \pm .01$        & $\mathbf{.17 \pm .0}$          &     $\mathbf{.03 \pm .01}$                
\end{tabular}
}
\end{table}

\subsection{Results} 
Experimental results are shown in Tables~\ref{exp:ihdp}, \ref{exp:twins} and \ref{exp:mimiciii}. 
%ABCEI performs stably well under different settings, with number of samples from $747$, $3212$ to $12828$ and number of covariates from $25$, $7$ to $43$.
%For IHDP dataset, as we can see, in the in-sample part ABCEI achieves a competitive result, which is very close to best method (SITE) on $\sqrt{\epsilon_{PEHE}}$. ABCEI has the best performance on the in-sample ATE, out-of-sample $\sqrt{\epsilon_{PEHE}}$ and ATE. 
%For Jobs dataset, ABCEI achieves better performance than other baselines on policy risk estimation and close performance to OLS/${LR_1}$ on in-sample ATT estimation. For Twins dataset, ABCEI has best performance outperforming other baselines.
%
It would be unsound to report statistical test results over the results reported in these tables; due to varying (un-)availability of ground truth, we must resort to reporting varying evaluation measures per dataset, over which it would not be appropriate to aggregate in a single statistical hypothesis test. However, one can see that ABCEI performs best in ten out of twelve cases, not only by the best number in the column, but often also by a non-overlapping empirical confidence interval with that of the best competitor (cf.\@ reported standard deviations).  This provides evidence that ABCEI is a substantial improvement over the state of the art.

% \begin{wraptable}{r}{0.69\textwidth}
\begin{table}
\caption{In-sample and out-of-sample results with mean and standard errors on the Twins dataset (AUC: higher = better, $\epsilon_{ATE}$: lower = better).}
\label{exp:twins}
\centering
% \resizebox{0.7\textwidth}{!}
{%
\begin{tabular}{l|l|l|l|l}
\multirow{2}{*}{Methods} & \multicolumn{2}{c|}{In-sample} & \multicolumn{2}{c}{Out-sample}                     \\ \cline{2-5} 
                         & \multicolumn{1}{c|}{$AUC$}        & \multicolumn{1}{c|}{$\epsilon_{ATE}$}       & \multicolumn{1}{c|}{$AUC$} & $\epsilon_{ATE}$ \\ \hline
OLS/$LR_1$ & $.660 \pm .005$            & $.004 \pm .003$                 & $.500 \pm .028$          &     $.007 \pm .006$       \\
OLS/$LR_2$ & $.660 \pm .004$                 & $.004 \pm .003$                 & $.500 \pm .016$          &     $.007 \pm .006$       \\
BLR        & $.611 \pm .009$                 & $.006 \pm .004$                 & $.510 \pm .018$          &     $.033 \pm .009$       \\
BART       & $.506 \pm .014$                 & $.121 \pm .024$                 & $.500 \pm .011$          &     $.127 \pm .024$       \\ \hline
k-NN       & $.609 \pm .010$                 & $.003 \pm .002$                 & $.492 \pm .012$          &     $.005 \pm .004$       \\ \hline
BNN        & $.690 \pm .008$                & $.006 \pm .003$                & $.676 \pm .008$          &     $.020 \pm .007$       \\
TARNet     & $.849 \pm .002$                 & $.011 \pm .002$                 & $.840 \pm .006$          &     $.015 \pm .002$       \\
CFR-Wass   & $.850 \pm .002$                 & $.011 \pm .002$                 & $.842 \pm .005$          &      $.028 \pm .003$      \\
CEVAE      & $.845 \pm .003$                 & $.022 \pm .002$                 & $.841 \pm .004$          &     $.032 \pm .003$       \\
SITE       & $.862 \pm .002$                 & $.016 \pm .001$                 & $.853 \pm .006$          &     $.020 \pm .002$       \\ \hline
$\text{ABCEI}^*$       & $.861 \pm .001$        & $.005 \pm .001$        & $.851 \pm .001$          &     $.006 \pm .001$ \\
$\text{ABCEI}^{**}$     & $.855 \pm .001$        & $.005 \pm .001$        & $.849 \pm .001$          &     $.006 \pm .001$  \\
$\text{ABCEI}$      & $\mathbf{.871 \pm .001}$        & $\mathbf{.003 \pm .001}$        & $\mathbf{.863 \pm .001}$          &     $\mathbf{.005 \pm .001}$     
\end{tabular}
}
\end{table}
% \end{wraptable}

\begin{table}[t]
\caption{In-sample and out-of-sample results with mean and standard errors on the MIMICIII benchmark (lower = better).}
\label{exp:mimiciii}
\centering
% \begin{tabular}{l|c|c|cc}
% \resizebox{0.98\textwidth}{!}
{%
\begin{tabular}{l|l|l|l|l}
{\multirow{2}{*}{Methods}} & \multicolumn{2}{c|}{In-sample} & \multicolumn{2}{c}{Out-sample} \\ \cline{2-5}
% \multirow{2}{*}{Methods} & \multicolumn{2}{c|}{In-sample} & \multicolumn{2}{c}{Out-sample}                     \\ \cline{2-5} 
{} & $\sqrt{\epsilon_{PEHE}}$ & $\epsilon_{ATE}$ & $\sqrt{\epsilon_{PEHE}}$ & $\epsilon_{ATE}$    \\ \hline
OLS/$LR_1$ & $7.1\phantom{0} \pm .2$  & $.92 \pm .15$ & $8.2\phantom{0} \pm .2$  & $.97 \pm .15$  \\
OLS/$LR_2$ & $2.7\phantom{0} \pm .1$  & $.24 \pm .11$ & $3.3\phantom{0} \pm .2$  & $.29 \pm .13$  \\
BLR        & $7.3\phantom{0} \pm .1$  & $.90 \pm .09$ & $8.5\phantom{0} \pm .3$  & $.97 \pm .09$  \\
BART       & $2.4\phantom{0} \pm .2$  & $.31 \pm .09$ & $3.1\phantom{0} \pm .2$  & $.37 \pm .12$  \\ \hline
k-NN       & $2.8\phantom{0} \pm .1$  & $.32 \pm .11$ & $3.6\phantom{0} \pm .1$  & $.36 \pm .11$  \\
RF         & $4.6\phantom{0} \pm .3$  & $.88 \pm .10$ & $5.3\phantom{0} \pm .3$  & $.89 \pm .11$  \\
CF         & $4.1\phantom{0} \pm .1$  & $.22 \pm .13$ & $4.9\phantom{0} \pm .1$  & $.24 \pm .14$  \\ \hline
BNN        & $2.5\phantom{0} \pm .1$  & $.45 \pm .11$ & $3.3\phantom{0} \pm .1$  & $.49 \pm .11$  \\
TARNet     & $1.91 \pm .0$  & $.25 \pm .16$ & $2.11 \pm .1$  & $.31 \pm .16$  \\
CFR-Wass   & $1.06 \pm .0$  & $.19 \pm .14$ & $1.09 \pm .0$  & $.21 \pm .14$  \\
CEVAE      & $2.71 \pm .0$  & $.23 \pm .11$ & $2.72 \pm .0$  & $.23 \pm .12$  \\
SITE       & $1.29 \pm .0$  & $.21 \pm .14$  & $1.35 \pm .0$   & $.25 \pm .14$ \\ \hline
$\text{ABCEI}^*$  & $\phantom{0}.89 \pm .0$ & $.13 \pm .13$  & $\phantom{0}.92 \pm .0$  &  $.16 \pm .14$  \\
$\text{ABCEI}^{**}$  & $\phantom{0}.96 \pm .0$ & $.15 \pm .12$  & $\phantom{0}.99 \pm .0$ &     $.16 \pm .14$ \\
$\text{ABCEI}$      & $\mathbf{\phantom{0}.85 \pm .0}$  & $\mathbf{.11 \pm .12}$      & $\mathbf{\phantom{0}.89 \pm .0}$          &     $\mathbf{.12 \pm .14}$                
\end{tabular}
}
\end{table}

Due to the existence of treatment selection bias, regression based methods suffer from high generalization error. Nearest neighbor based methods consider 
%the similar units 
unit similarity 
to overcome selection bias, but cannot achieve balance globally.  
Recent advances in representation learning bring improvements in causal effect estimation. 
% CEVAE uses variational autoencoders to learn the latent variable causal model, which has competitive performance on the Jobs and Twins datasets, but relatively weak performance on the IHDP dataset.
% BNN accounts for selection bias balancing, but considers treatment variable as a feature using one single network, which might lose the impact of treatment variable when the dimension of representations is high~\cite{shalit2017estimating}. TARnet employs two separate networks but follows agnostic setting without balancing selection bias. 
% CFR-Wass does not consider local similarity information in the original covariate space. SITE uses hard samples to preserve local similarity information and consider balance property. 
Unlike CFR-Wass, BNN, and SITE, ABCEI considers information loss and balancing problems. The mutual information estimator ensures that the encoder learns representations preserving useful information from the original covariate space. The adversarial learning component constrains the encoder to learn balanced representations. This causes ABCEI to achieve better performance than the baselines. We also report the performance of our model without mutual information estimator or adversarial learning, respectively, as $\text{ABCEI}^*$, $\text{ABCEI}^{**}$. From the results we can see that performance suffers when either of these components is left out,
%both of their performance are worse than the original model, 
which demonstrates the importance of combining adversarial learning and mutual information estimation in ABCEI. 

\begin{figure}[t]
  \centering
  \includegraphics[width=0.7\columnwidth]{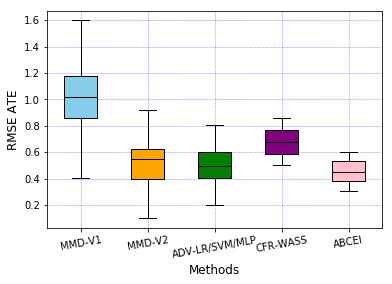}
  \caption{Results on ACIC datasets.}
  \label{fig:acic}
\end{figure}

In Figure~\ref{fig:acic}, we compare ABCEI with recent balancing methods on ACIC benchmark. As we can see, the variance of representation learning methods are lower than methods reweighing samples on covariate space. We also found that the adversarial balancing methods perform better on ATE estimation. ABCEI has the advantage of adversarial balancing as well as preserving predictive information in latent space, which makes it outperforms the other baselines. 

% \begin{figure*}[t]
%   \centering
%   \subfloat[\label{fig:KLdv}]{\includegraphics[width=0.43\columnwidth]{{KLdv}}}
%   \hspace{1.5mm}
%   \subfloat[\label{fig:gmi_pehe}]{\includegraphics[width=0.47\columnwidth]{gmi_pehe}}
%   \caption{Empirical robustness analysis.}
% \end{figure*}

\subsection{Training details}

We adopt ELU~\citep{clevert2015fast} as the non-linear activation function if there is no specification.
We employ various numbers of fully-connected hidden layers with various sizes across networks: four layers with size $200$ for the encoder network; two layers with size $200$ for the mutual information estimator network; three layers with size $200$ for the discriminator network; and finally, three layers with size $100$ for the predictor network, following the structure of TARnet~\citep{shalit2017estimating}. The gradient penalty weight $\beta$ is set to $10.0$, and the regularization weight is set to $0.0001$. 
% Adversarial learning process is shown in Figure~\ref{fig:ganbal}.

In the training step, firstly we minimize $L_{\Phi \Omega}$ by simultaneously optimizing $\Phi$ and $\Omega$ with one-step gradient descent. Then the representations $h$ are passed to the discriminator to minimize $L_D$ by optimizing $D$ with $3$-step gradient descent, in order to find a stable discriminator. Next, we use discriminator $D$ to train encoder $\Phi$ by minimizing $L_{\Phi}$ with one-step gradient descent. Finally, encoder $\Phi$ and predictor $\Psi$ are optimized simultaneously by minimizing $L_{\Phi \Psi}$.

% \begin{figure*}[t]
%   \centering
%   \includegraphics[width=\columnwidth]{GAN_rep}
%   \caption{Adversarial learning structure for representation balancing.}
%   \label{fig:ganbal}
% \end{figure*}

\subsection{Hyper-parameter optimization}

Due to the reason that we cannot observe counterfactuals in observational datasets, standard cross-validation methods are not feasible. We follow the hyper-parameter optimization criterion in~\citep{shalit2017estimating}, with an early stopping with regard to the lower bound on the validation set. Detail search space of hyper-parameter is demonstrated in Table~\ref{tb:hp_optimize}. The optimal hyper-parameter settings for each benchmark dataset is demonstrated in Table~\ref{tb:hp_optimal}.

\begin{table}[htb]
\caption{Search space of hyper-parameter}
\label{tb:hp_optimize}
\centering
\begin{tabular}{l|l}
Hyper-parameter & Range \\ \hline
$\lambda$ & $1e$-$3$,$1e$-$4$,$5e$-$5$  \\ \hline
$\beta$ & $1.0$,$5.0$,$10.0$,$15.0$ \\ \hline
Optimizer & RMSProp, Adam \\ \hline
Depth of encoder layers& $1,2,3,4,5,6$ \\ \hline
Depth of discriminator layers & $1,2,3,4,5,6$ \\ \hline
Depth of predictor layers & $1,2,3,4,5,6$ \\ \hline
Dimension of encoder layers & $50,100,200,300,500$ \\ \hline
Dimension of discriminator layers & $50,100,200,300,500$ \\ \hline
Dimension of MI estimator layers & $50,100,200,300,500$ \\ \hline
Dimension of predictor layers & $50,100,200,300,500$ \\ \hline
Batch size & $65,80,100,200,300,500$
\end{tabular}
\end{table}

% Please add the following required packages to your document preamble:
% \usepackage{multirow}
\begin{table}[htb]
\caption{Optimal hyper-parameter for each benchmark dataset}
\label{tb:hp_optimal}
\centering
\begin{tabular}{l|cccc}
\multirow{2}{*}{Hyper-parameters} & \multicolumn{4}{c}{Datasets} \\ \cline{2-5} 
 & \multicolumn{1}{c|}{IHDP} & \multicolumn{1}{c|}{Jobs} & \multicolumn{1}{c|}{Twins} & ACIC \\ \hline
$\lambda$ & \multicolumn{1}{c|}{$1e$-$4$} & \multicolumn{1}{c|}{$1e$-$4$} & \multicolumn{1}{c|}{$1e$-$4$} & $1e$-$4$ \\ \hline
$\beta$ & \multicolumn{1}{c|}{$10.0$} & \multicolumn{1}{c|}{$10.0$} & \multicolumn{1}{c|}{$10.0$} & $10.0$ \\ \hline
Optimizer & \multicolumn{1}{c|}{Adam} & \multicolumn{1}{c|}{Adam} & \multicolumn{1}{c|}{Adam} & Adam \\ \hline
Depth of encoder layers & \multicolumn{1}{c|}{$4$} & \multicolumn{1}{c|}{$5$} & \multicolumn{1}{c|}{$5$} & $4$ \\ \hline
Depth of discriminator layers & \multicolumn{1}{c|}{$3$} & \multicolumn{1}{c|}{$3$} & \multicolumn{1}{c|}{$3$} & $3$ \\ \hline
Depth of predictor layers & \multicolumn{1}{c|}{$3$} & \multicolumn{1}{c|}{$3$} & \multicolumn{1}{c|}{$3$} & $3$ \\ \hline
Dimension of encoder layers & \multicolumn{1}{c|}{$200$} & \multicolumn{1}{c|}{$200$} & \multicolumn{1}{c|}{$300$} & $200$ \\  \hline
Dimension of discriminator layers & \multicolumn{1}{c|}{$200$} & \multicolumn{1}{c|}{$200$} & \multicolumn{1}{c|}{$200$} & $200$ \\ \hline
Dimension of MI estimator layers & \multicolumn{1}{c|}{$200$} & \multicolumn{1}{c|}{$200$} & \multicolumn{1}{c|}{$200$} & $200$ \\ \hline
Dimension of predictor layers & \multicolumn{1}{c|}{$100$} & \multicolumn{1}{c|}{$100$} & \multicolumn{1}{c|}{$200$} & $100$ \\ \hline
Batch size & \multicolumn{1}{c|}{$65$} & \multicolumn{1}{c|}{$100$} &  \multicolumn{1}{c|}{$300$} & $100$ 
\end{tabular}
\end{table}

\subsection{Computational complexity}
Assuming the size of mini-batch is $n$, number of epochs is $m$, the computational complexity of our model is $\mathcal{O}(n \ast m \ast ((\Phi_h-1)\Phi_w^2 + (\Omega_h-1)\Omega_w^2 + (D_h-1)D_w^2 + (\Psi_h-1)\Psi_w^2))$. Here $\Phi_h, \Omega_h, D_h, \Psi_h$ indicates the number of layers and $\Phi_w, \Omega_w, D_w, \Psi_w$ indicates number of neurons in each layer in Neural Network $\Phi, \Omega, D, \Psi$.

\subsection{Robustness Analysis on Selection Bias}
To investigate the performance of our model when varying the level of selection bias, we generate toy datasets by varying the discrepancy between the treatment and control groups. We draw $8\thinspace000$ samples with ten covariates $x \sim \mathcal{N}(\mu_0, 0.5 \cdot (\Sigma + \Sigma^T))$ as control group, where $\Sigma \sim \mathcal{U}((-1,1)^{10 \times 10})$. Then we draw $2\thinspace000$ samples from $x \sim \mathcal{N}(\mu_1, 0.5 \cdot (\Sigma + \Sigma^T))$. By adjusting $\mu_1$, we generate treatment groups with varying selection bias, which can be measured by KL-divergence. For the outcomes, we generate $y|x \sim (w^Tx + n)$, where $n \sim \mathcal{N}(0^{2 \times 1}, 0.1 \cdot I^{2 \times 2})$ and $w \sim \mathcal{U}((-1,1)^{10 \times 2})$. 

\begin{figure}[t]
  \centering
  \includegraphics[width=0.7\columnwidth]{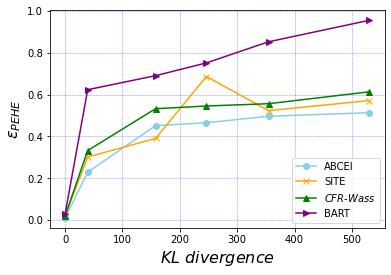}
  \caption{$\epsilon_{PEHE}$ on datasets with varying treatment selection bias. ABCEI is comparatively robust.}
  \label{fig:KLdv}
\end{figure}

In Figure~\ref{fig:KLdv}, we can see the robustness of ABCEI, in comparison with CFR-Wass, BART, and SITE. The reported experimental results are averaged over 100 test sets. From the figure, we can see that with increasing KL-divergence, our method achieves more stable performance. We do not visualize standard deviations as they are negligibly small.

\begin{figure}[t]
  \centering
  \includegraphics[width=0.7\columnwidth]{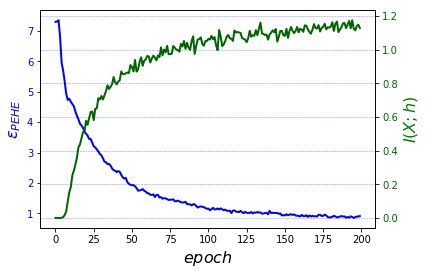}
  \caption{Mutual information (MI) between representations and original covariates, as well as $\epsilon_{PEHE}$ in each epoch. With increasing MI, $\epsilon_{PEHE}$ decreases.}
  \label{fig:gmi_pehe}
\end{figure}

\subsection{Robustness Analysis on Mutual Information Estimation} 
To investigate the impact of minimizing the information loss on causal effect learning, we block the adversarial learning component and train our model on the IHDP dataset. We record the values of the estimated MI and $\epsilon_{PEHE}$ in each epoch. In Figure~\ref{fig:gmi_pehe}, we report the experimental results averaged over 1000 test sets. We can see that with increasing MI, the mean square error decreases and reaches a stable region. But without the adversarial balancing component, the $\epsilon_{PEHE}$ cannot be further lowered due to the selection bias. This result indicates that even though the estimators benefit from highly predictive information, they will still suffer if imbalance is ignored. 

\subsection{Balancing Performance of Adversarial Learning}
\begin{figure*}[t]
  \centering
  \subfloat[IHDP.\label{fig:ihdp}]{\includegraphics[width=\columnwidth]{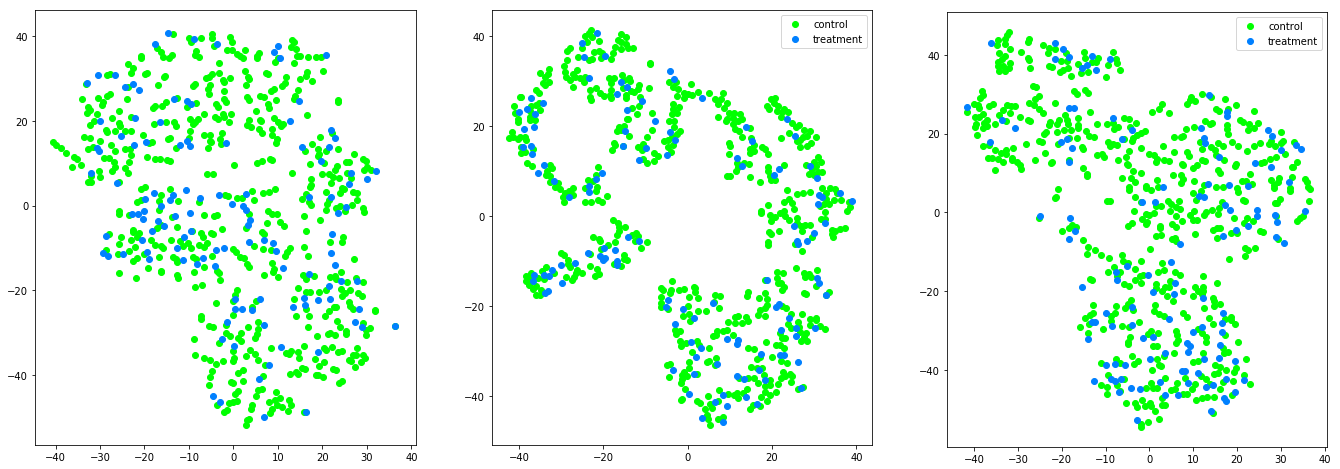}}
  \vspace{1mm}
  \subfloat[Jobs.\label{fig:jobs}]{\includegraphics[width=\columnwidth]{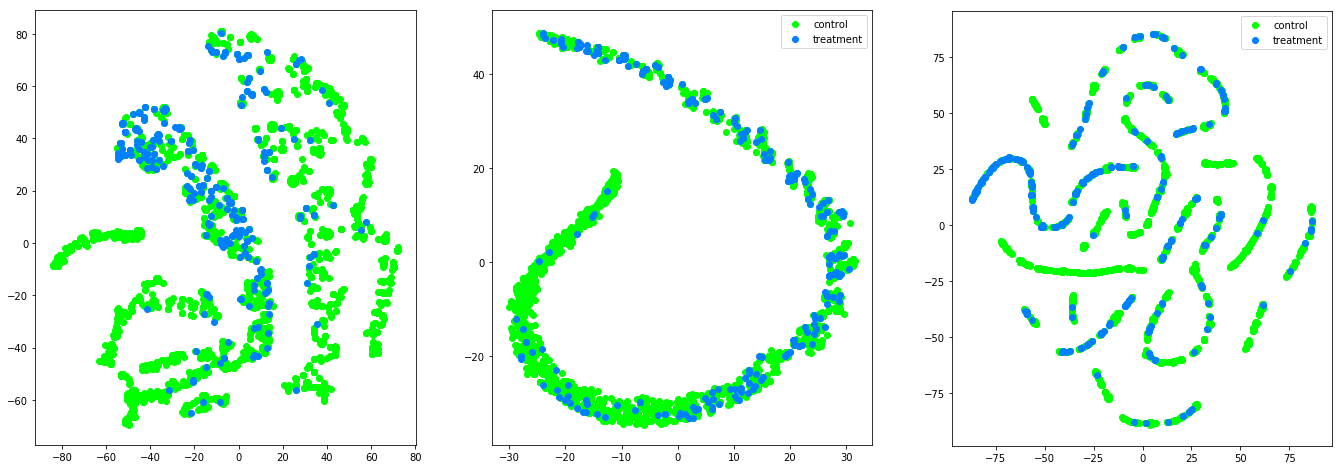}}
  \caption{t-SNE visualization of treatment and control group, on the IHDP and Jobs datasets. The blue dots are treated units, and the green dots are control units. The left figures are the units in original covariate space, the middle figures are representations learned by ABCEI, and the right figures are representations learned by CFR-Wass; notice how the latter has control unit clusters unbalanced by treatment observations.}
  \label{fig:hrep_clusters}
\end{figure*}

In Figure~\ref{fig:hrep_clusters}, we visualize the learned representations on the IHDP and Jobs datasets using t-SNE. We can see that compared to CFR-Wass, the coverage of the treatment group over the control group in the representation space learned by our method is better. 
This showcases the degree to which adversarial balancing improves the performance of ABCEI, especially in population causal effect (ATE, ATT) inference.

\section{Related Work}
Studies on causal effect inference give us insight on the true data generating process and allow us to answer the what-if questions. The core issue of causal effect inference is the identifiability problem given some data and set of assumptions~\citep{tian2002general}. Such data includes experimental data from Randomized Controlled Trials (RCTs) and non-experimental data that are collected from historic observations. 
Due to the difficulties of conducting RCTs, we mainly focus on the study of causal effect inference based on observational data.
There are many research focus on the identifiability of determining cause from effect~\citep{mooij2016distinguishing,marx2019identifiability}. 

In this paper, we focus on the study of assessing the strength of causal effect with the assumptions of causal relations. Confounding bias might create spurious correlations between variables and would lead to difficulties for the identification of causal effect with observational data. Strong ignorability assumption in Potential Outcome framework~\citep{rubin2005causal} provides a way to remove the confounding bias and make causal effect inference possible with observational data. For practical applications, there are some studies focusing on matching based methods~\citep{ho2011matchit} to create comparable groups for causal effect inference. Various similarity measures are applied to achieve better matching results and reduce the estimation error, e.g. Mahalanobis distance and propensity score matching methods are proposed for population causal effect inference~\citep{rubin2001using,diamond2013genetic}. An information theory-driven approach is proposed by using mutual information as the similarity measure~\citep{sun2016mutual}. 

Recent studies employ deep representation learning methods to derive models that satisfy the conditional ignorability assumption~\citep{li2017matching}, so that make the Conditional Average Treatment Effect identifiable, e.g. ~\cite{johansson2016learning} propose to use a single neural network with the concatenation of representations and treatment variable as the input to predict the potential outcomes. ~\cite{shalit2017estimating} propose to train separate models for different treatment outcome systems associating with a measure based on probabilistic integral metric to bound the generalization error.~\cite{yao2018representation} propose to employ hard samples to preserve local similarity in order to achieve better balancing results. The main difference between ABCEI and the state-of-the-art representation learning based methods are two-fold: on the one hand, by employing adversarial learning, our balancing method does not need any assumptions on the treatment selection functions; on the other hand, the transformation between original covariate space to the latent space might lead to information loss. In our framework, a mutual information estimator is employed to enforce the encoder preserving as much as highly predictive information.

From the view of graphical interpretation, there are some other difficulties for the identification of causal effect, e.g. selection bias~\citep{correa2019identification}. Some research~\citep{bareinboim2012controlling} propose the use instrumental variable for the identification of causal effect. In this paper, we assume there exists only confounding bias, so that removing confounding bias can make the causal effect identifiable.
Due to the unjustifiable property of strong ignorability, controlling covariates may not remove confounding bias when there exists unobserved confounders. Some research propose to estimate causal effect by using proxy variables~\citep{louizos2017causal}. A modified variational autoencoder structure is employed to identify the causal effect from observational data. In this paper, we assume that all the confounder can be measured, so that our method is sufficient for the identifiability of the CATE.

\section{Conclusions}

We propose a novel model for causal effect inference with observational data, called \emph{ABCEI}, which is built on deep representation learning methods. ABCEI focuses on balancing latent representations from treatment and control groups by designing a two-player adversarial game. We use a discriminator to distinguish the representations from different groups. By adjusting the encoder parameters, our aim is to find an encoder that can fool the discriminator, which ensures that the distributions of treatment and control representations are as similar as possible. Our balancing method does not make any assumption on the form of the treatment selection function. 
% In this model, we engage two important performance-ensuring components: mutual information estimator and adversarial balancing. These two components can help us to tackle the problems of information loss and selection bias in designing the causal effect predictors. 
%On the other hand, w
With the mutual information estimator, we preserve highly predictive information from the original covariate space to latent space. 
Experimental results on benchmark datasets and synthetic datasets demonstrate that ABCEI is able to achieve robust, and substantially better performance than the state of the art.

% Experimental results demonstrate the effectiveness of mutual information on causal effect estimation. 
% At the same time, adversarial learning balances the representation distributions of treatment and control group. Experimental results show that our encoder can learn more identical distributions of treatment and control groups with better coverage. Finally, 
In future work, we will explore more connections between relevant methods in domain adaptation~\citep{daume2006domain} and counterfactual learning~\citep{swaminathan2015counterfactual} with the methods in causal inference. A proper extension would be to consider multiple treatment assignments or the existence of hidden confounders. 
% Furthermore, we also plan to investigate the causal effects in subpopulations to detect the latent heterogeneity, which is a very important issue for decision makers in many fields such as public health and social security.

% \bibliographystyle{spbasic}
% \bibliography{causal}

% \appendix

% \section{Proofs}

% \subsection{Donsker-Varadhan}
% \begin{theorem}[Donsker-Varadhan]
% Let $P$, $Q$, $G$ be distributions on the same support $\mathcal{Z}$, and let $\mathcal{C}$ denote a family of functions $\Omega: \mathcal{Z} \rightarrow \mathbb{R}$, we have
% \begin{equation*}
% \begin{split}
% &D_{KL}(P||Q) = \sup_{\Omega \in \mathcal{C}} \mathbb{E}_{P}[\Omega(Z)] - \log\mathbb{E}_{Q}\left[e^{\Omega}(Z)\right]
% \end{split}
% \end{equation*}
% \end{theorem}

\end{document}